\def\year{2022}\relax
\documentclass[letterpaper]{article} % DO NOT CHANGE THIS
\usepackage{aaai21}  % DO NOT CHANGE THIS
\usepackage{times}  % DO NOT CHANGE THIS
\usepackage{helvet} % DO NOT CHANGE THIS
\usepackage{courier}  % DO NOT CHANGE THIS
\usepackage[hyphens]{url}  % DO NOT CHANGE THIS
\usepackage{graphicx} % DO NOT CHANGE THIS
\usepackage[table]{xcolor}
\usepackage{natbib}  % DO NOT CHANGE THIS AND DO NOT ADD ANY OPTIONS TO IT
\usepackage{caption} % DO NOT CHANGE THIS AND DO NOT ADD ANY OPTIONS TO IT
\usepackage{todonotes}
\usepackage[toc]{appendix}
\usepackage[linesnumbered, ruled, noend]{algorithm2e}
\usepackage{setspace}%tighter spacing on algorithms in methodology

\usepackage{amsmath}
\title{Provably Robust Model-Centric Explanations for Critical Decision-Making}
\author{
    Cecilia G. Morales,
    Nicholas Gisolfi,
    Robert Edman,
    James K. Miller, 
    Artur Dubrawski
    \\
}
\affiliations{
    Auton Lab, Carnegie Mellon University,
    5000 Forbes Avenue, Pittsburgh, PA 15213\\
    \{cgmorale,ngisolfi,redman,mille856,awd\}@andrew.cmu.edu

}
\begin{document}

\maketitle

\begin{abstract}
% V2
We recommend using a model-centric, Boolean Satisfiability (SAT) formalism to obtain useful %minimal distance counterfactual 
explanations of trained model behavior, different and complementary to 
what can be gleaned from LIME and SHAP, popular data-centric explanation tools in Artificial Intelligence (AI).
We compare and contrast these methods, and show that data-centric methods may yield brittle explanations of limited practical utility. %, in the sense that they change when different data is under consideration.
The model-centric framework, however, can offer actionable insights into risks of using AI models in practice. % differs in that the model is static, therefore a single explanation exists.
For critical applications of AI, split-second decision making is best informed by robust explanations that are invariant to properties of data, the capability offered by model-centric frameworks. 

\end{abstract}

\section{Introduction}
Artificial Intelligence (AI)-driven decision making is increasingly used to support human decisions. 
In practice, the adoption of intelligent systems hinges on the ability of users to understand why a prediction was generated and how to ensure a desired output. 
Trust and transparency in AI is essential in high-stakes application domains, including healthcare, counter-terrorism, management of nuclear facilities, etc., where wrong decisions may bear grave consequences. 
Lack of trust in AI systems is often due to lack of their understanding or interpretability ~\cite{10.1145/3239060.3239064}, especially when safety is at risk ~\cite{10.1145/3322276.3322345}.

Common explanatory tools, including Local Interpretable Model-Agnostic Explanations (LIME) ~\cite{lime} and Shapley Additive Explanations (SHAP) ~\cite{NIPS2017_7062},  are data-centric. 
They assess contributions of individual attribute values to predictive performance of the models. Insights gleaned from such analyses are primarily of confirmatory value; a clinician can confirm that the model pays attention to similar features that she would consider in analyzing her current patient. 
However, these tools can be brittle, hide biases,
and do not provide useful diagnostic information about safety of the models.

We define brittleness of a model by its inability to adapt or generalize to conditions outside of a narrow set of assumptions. Brittleness of a model could be lethal in safety-critical domains since an anomalous data point could be classified incorrectly. Further decreasing trust in the models. Alternatively, the explanations given by these models are of feature importance and do not inform the user what would be needed to change the outcome of the model, also dissuading people from using them. We define feature importance in the rest of the paper as the measure of an individual contribution of the corresponding feature for a the classification performance of the model~\cite{article}.\par

Conversely, formal methods can be used to mathematically prove desired reliability properties of the models, eliminating human biases and statistical errors from the process~\cite{gisolfi2021formal}.

We extend these methods to provide minimal-distance counterfactuals that find minimal changes to attribute values needed to cause the model to change its prediction.
This analysis can expose limitations of models and data used to train them, enabling development of provably robust AI-driven decision support systems.

%This mix of provable guarantees along with the capability to explain the model directly, rather than explaining model behavior with appeals to related data points, increase the trustworthiness of a model.

\section{Related Work}

Model interpretability is currently used to bridge the lack of trust in models in machine learning. It can be approached by model: intrinsic or post-hoc, by scope: local or global, or by method: model-specific or model-agnostic. They have the common goal to present concise and intelligible information to a human user that empowers them to determine whether or not the system is behaving in a safe manner. LIME and SHAP are among the most popular explainable methods. Both are post-hoc, local and model-agnostic. Nevertheless they come with a variety of drawbacks. Some of the disadvantages of these methods are that they require a human to perform a verification task to certify if the model behaves as expected, nonetheless, the process is error prone. Moreover, post-hoc explanation techniques might be unstable and unreliable~\cite{10.1145/3236386.3241340}. Ghorbani et al. show that some explanation techniques can be highly sensitive to small perturbations in the input even though the underlying classifier's predictions remain unchanged~\cite{ghorbani2019interpretation}.\par 

The LIME method interprets individual model predictions based on linear assumptions and locally approximating the model around a given prediction but the correct definition of the neighborhood is unknown~\cite{lime}. LIME requires of human input to determine the correct kernel setting based on whether the explanations are coherent; however, this process is susceptible to mistakes ~\cite{molnar2019}. Alvarez-Melis and Jaakkola showed that the explanations were very unstable given that for two very close points they varied greatly in a simulated setting and were often inconsistent with each other ~\cite{AlvarezMelis2018OnTR}. \par

The SHAP method is derived from Shapley values in game theory, where
they describe the contribution of each team player in a collaborative environment. In machine learning, it works by assigning each feature an importance value for a particular prediction~\cite{NIPS2017_7062}. One of the distinguishing factors of SHAP is that the prediction is fairly distributed among features. However some disadvantages according to Molnar ~\cite{molnar2019} are that users prefer selective explanations since an overwhelming number of features could create confusion among users. Since Shapley values return a value per feature, no conclusion can be made regarding a change in prediction for changes in the input. Similarly to many permutation-based interpretation methods, the Shapley value method can't classify unrealistic data instances when features are correlated. This works however if features are independent since when a feature is missing it is then marginalized. This is achieved by sampling values from the feature's marginal distribution. \par

An explanation can be viewed from a constraint satisfaction problem perspective; can a model satisfy all constraints, and if not, why not? Satisfiability-based formal methods, i.e. SAT, SMT, MILP, provide these types of proofs for abstract mathematical representation of trained models. The explanations are instances of constraint satisfaction problems. Formal methods are on increasing interest in providing explanations for model behaviors~\cite{karimi2020model}. Formal methods have also been used to assess the quality of explainable AI methods ~\cite{narodytska2019assessing}. 

Providing explanations for model behavior under ideal conditions is still a significant challenge, however, one could argue that there is even higher imperative to provide explanations for model behavior under tough conditions. For instance, providing an explanation that illustrates a similarity between model predictions on two similar samples of data will not suffice when one of the data points in question is anomalous and no sufficiently similar neighbors exist. Explanations like the ones provided by LIME and SHAP are too brittle and thus should not be used for safety-critical applications.

\section{Methodology}
\textbf{Data and Models.}
We use the publicly available Breast Cancer Wisconsin (Diagnostic) dataset~\cite{Dua:2019}.
It contains 30 numeric features which we standardized by removing the mean and scaling to unit variance. We explain Scikit-learn~\cite{scikit-learn} 
random forests, trained with a 50\% train/test split. Our model consists of ten decision trees of the maximum depth of ten.\par

 \begin{figure}
     \centering
     \includegraphics[width=\columnwidth, height = 2cm]{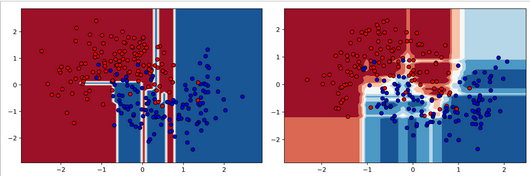}
     \caption{Decision Surface of a Decision Tree [left] and Random Forest [right]}
     \label{fig:variable_cdf}
 \end{figure}

\begin{figure}
    \centering
    \includegraphics[width=\columnwidth, height = 3cm]{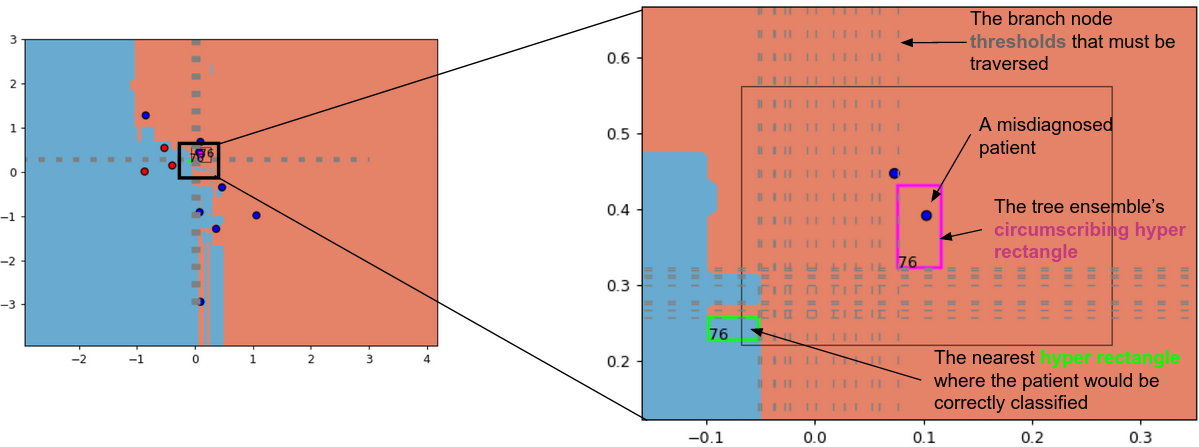}
    \caption{Illustration of a minimal distance counterfactual.  A sample in the magenta box will receive a different label if the sample had attribute values that placed it inside the green box.  Our SAT framework can produce these explanations regardless of whether there is supporting data in the region under consideration.}
    \label{fig:variable_cdf}
\end{figure}

\noindent
\textbf{Experiments.}
We first created explanations for the same data point following the methods of LIME, SHAP and SAT, which are described in detail next. 
LIME ~\cite{lime} provides local explanations of model predictions global by creating an approximation near the input. Feature importance can be inferred from this local linear approximation by looking at the feature weights. For our experiments, we mirrored the feature importance experiments of the LIME repository\footnote{https://github.com/marcotcr/lime}. 
%understanding of the model by explaining a set of individual instances that are sampled by a Gaussian distribution.
% Its approach is similar to computing summary statistics such as held-out accuracy. Let $\textit{g} \in \textit{G}$, where \textit{G} is a class of potentially interpretable models. The explanation produced by LIME is the following: 
% \begin{equation}
% \xi (x) = \argmin\limits_{g\in G}  \mathcal{L} (f, g, \pi_{x}) + \Omega(g)
% \end{equation}
% Where $\mathcal{L} (f, g, \pi_{x})$ is minimized which is a measure of how unfaithful a model \textit{g} is in approximating \textit{f} in the locality defined by $\pi_{x}$ and $\Omega(g)$ is a measure of complexity of the explanation $\textit{g} \in \textit{G}$. 
%The LIME explanations were generated from  
\begin{figure*}
    \centering
    \includegraphics[width=\textwidth]{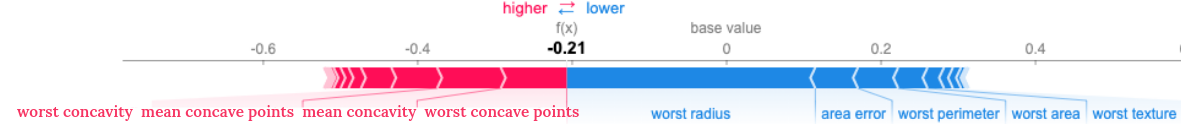}
    \caption{SHAP Explanation- Force Plot}
   \label{fig:shap_exp}

\end{figure*}

\begin{figure}
    \centering
    \includegraphics[width=\columnwidth, height=7cm]{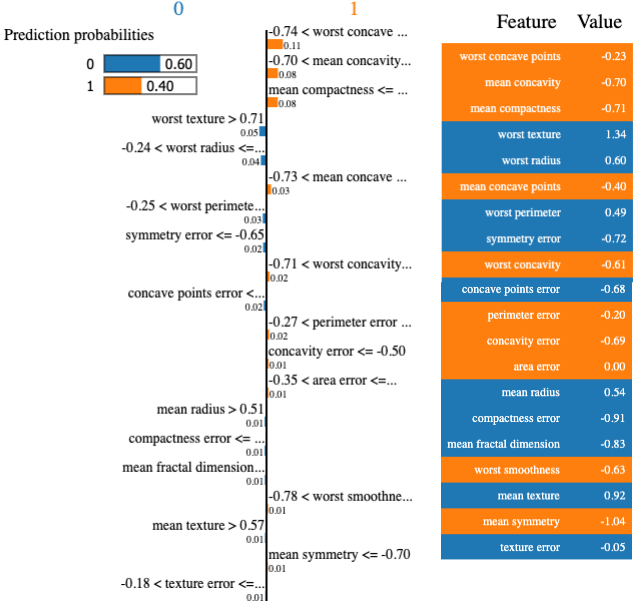}
    \caption{LIME Explanation - Feature Importance. Attributes are ranked in descending order of impact on the overall prediction probability yielded by the base model.}
   \label{fig:lime_exp}
\end{figure}
SHAP values ~\cite{NIPS2017_7062} provide the additive feature importance measure that adheres to local accuracy, missigness, and consistency. 
\begin{equation*}
    \phi_{\textit{i}} (\textit{f}, \textit{x}) = \sum_{z^\prime \subseteq x^\prime} \frac{\lvert z^\prime\rvert!(M -\lvert z^\prime\rvert- 1)!}{M!} [\textit{E}[\textit{f}(z)|\textit{z}_{s}] - \textit{f}(z^\prime\backslash i)]
\end{equation*}
Where M is the number of simplified input features, $\lvert z^\prime\rvert$ is the number of non-zero entries in $z^\prime$ and $z^\prime\subseteq x^\prime$ represents all $z^\prime$ vectors where the non-zero entries are a subset of the non-zero entries in $x^\prime$, and \textit{S} is the set of non-zero indexes in $z^\prime$. The SHAP explanations were generated from code by ~\cite{lundberg2018explainable}. 
\\
Our proposed framework uses the logical encoding strategy from~\cite{gisolfi2021formal}, and extends those methods to find the minimal distance counterfactual explanation as detailed in Algorithm \ref{alg:mindis}. To find minimal distance counterfactual explanations, we start with a data points and a local neighborhood in which to search for another data point to which the model assigns a different predicted label.  If such a point exists, a satisfying assignment detailing the model state for the two points is returned, otherwise, we increase the size of the local neighborhood and search again. This process continues until a counterfactual is found. The relevant differences between the two points which form the counterfactual can be revealed by finding all differences within the satisfying assignments. We only report encoded threshold values which must be crossed in order for the model to change its prediction, which may be interpreted as, 'if we change select attribute values in a query by a small amount, the model will change its prediction'. 
\begin{algorithm}
    \DontPrintSemicolon
    $\mathcal{M} \gets \text{trained random forest model}$\;
    $\mathbf{x} \gets \text{data point}$\;
    $\mathbf{\delta} \gets \text{initial local neighborhood}$\;
    $\phi \gets Encode(\mathcal{M},\mathbf{x},\mathbf{\delta})$\cite{gisolfi2021formal}\;
    \While{$\phi$ is \texttt{UNSAT}}{
        $\delta \gets \delta * 1.01$\;
        $\phi \gets Encode(\mathcal{M},\mathbf{x},\mathbf{\delta})$\;
    }
    $\mathbf{x}^\prime \gets \text{counterexample from SAT assignment of }\phi$\;
    \Return $\mathbf{x}^\prime - \mathbf{x} \text{, the counterfactual}$
    \caption{Minimal distance counterfactual}
    \label{alg:mindis}
\end{algorithm}

\begin{figure*}
    \centering
    \includegraphics[width=0.8\textwidth]{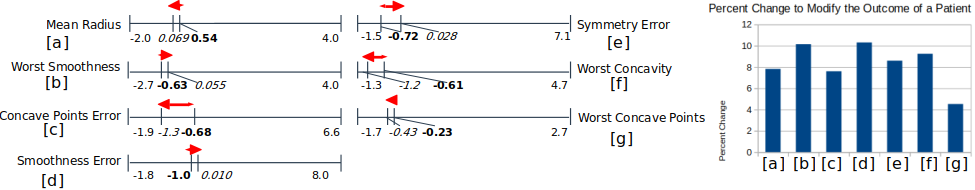}
    \caption{SAT Explanation - Counterfactual. Our explanation presents a set of the smallest changes necessary to change the model output. The bar graph shows the percent change that is required along each attribute to change the output of the model.}
    \label{fig:sat_exp}
\end{figure*}

We studied the stochasticity of feature importance in the LIME ~\cite{lime} and SHAP ~\cite{NIPS2017_7062} frameworks. Using those tools, we returned the $n$ most important features where $n = 1...30$. We iterated over all test data and recorded the $n$ most important features. We then evaluated the probability that each feature was one of the most influential variables in the test dataset when $n$ was set. We show feature importance attribution plots for LIME and SHAP where three features that were consistently picked as important for both are displayed in the bottom rows of Fig.~\ref{fig:variable_cdf} and the graphs for the rest of the features can be found in the Appendix figures \ref{fig:lime_summary} and \ref{fig:shap_summary}. The horizontal axis in each of these plots shows the index of the feature in the importance ranking and the vertical axis shows the estimated probability of the feature attaining such rank. Generally, truly important features would have this probability raise quickly as a function of the rank index, and stay high.  E.g., in the example shown in Fig.~~\ref{fig:variable_cdf}, feature \textit{mean concave points} appears slightly more important than \textit{worst concave points} and \textit{worst concavity}.

We explored explanations generated by SAT. Firstly, we looked at the 17 points in the test data that were misclassified. Each explanation had a set of features that needed to be changed to flip the prediction. We iterated over all the explanations and recorded the percentage change over the range of values of each feature that would be required to modify the prediction, and visualized their distributions in kernel density estimation plots. Then we repeated the same procedure with all the correctly classified points. The top row of Fig.~\ref{fig:variable_cdf} shows these characteristics for the same three features previously mentioned, the rest can be found in the Appendix figure \ref{fig:sat_summary}. Model-centric explanation suggests that fixing these 17 errors will require increasing the value of \textit{worst concavity} by 5-10\%. The other two features do not show such a consistent recipe for error correction, even though LIME suggests they are important.

\begin{figure}
    \centering
    \includegraphics[width=1.02\columnwidth]{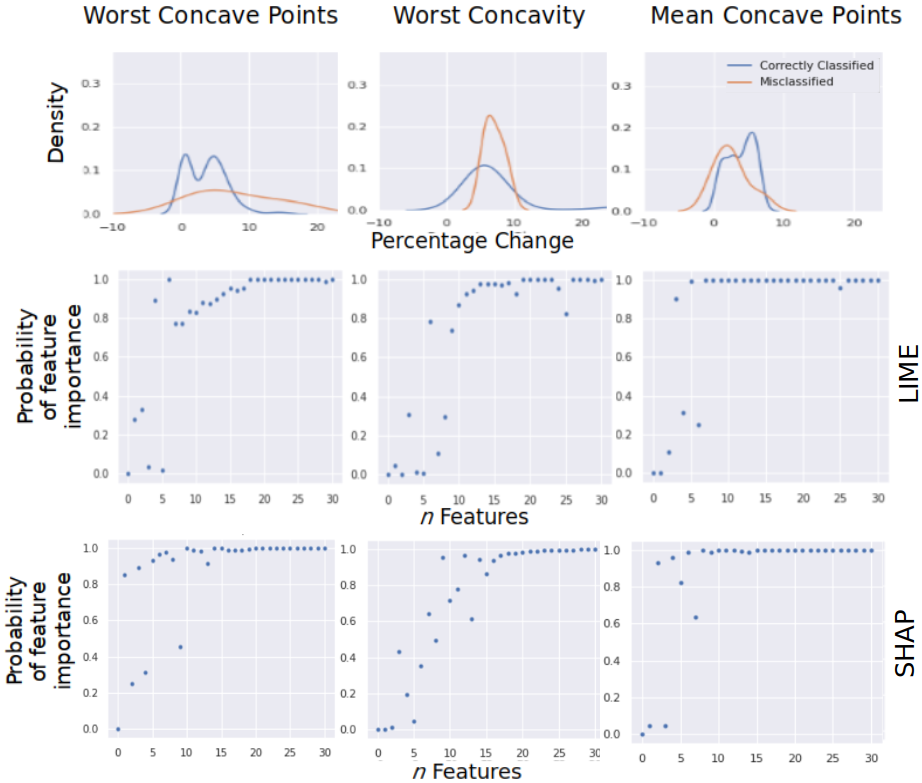}
    \caption{Aggregate summary of SAT counterfactual explanations [first] in addition to feature attribution for LIME [second] and SHAP [third].}
    \label{fig:variable_cdf}
\end{figure}

\section{Analysis }
The LIME explanations as shown in Figure ~\ref{fig:lime_exp} demonstrate feature attributions [left] where the most influential is at the top and current values [right] for the breast cancer mass. Feature importance attenuates quickly down the list, and attributes at the bottom have little relevance in the explanation at hand. An interesting observation is that the most relevant features in this particular explanation wind up describing the opposite of the predicted class. 

The SHAP features can be visualized as forces that influence the prediction, which starts at the baseline. The baseline value, in this case, \textit{-0.21}  represents the average of all predictions. The Shapley value is an arrow that pushes to increase, if positive, or decrease if negative as seen in Figure ~\ref{fig:shap_exp}.

Fig.~\ref{fig:sat_exp} shows an example model-centric explanation generated by our framework for one query. The current feature value for the suspected breast cancer mass is denoted in \textbf{bold} while the value it would need to be modified to is shown in \textit{italics}. The left part of the graph shows the seven attribute changes required by the model to change its prediction, and the right part shows the percentage change of value of each feature needed to cause such change. This type of explanation brings our attention to features that can compromise robustness of the model: the lower the relative value change needed to affect the output, the narrower the margin for measurement error in the model. 

This result has multiple potentially useful consequences. One example is confirmatory analysis of a prediction made by an AI-driven tool used by a clinician to help her diagnose a patient. If only a small change of one of the features, reflecting, e.g., some laboratory test result, can flip the prediction, the doctor may consider repeating the test to ascertain its outcome, or order a more precise test. 
Similarly, when designing AI-based decision systems that operate on measurements collected with noisy sensors with known sensor noise models, the design engineer may use the proposed analysis to verify that the expected magnitude of sensor noise does not exceed the range of robustness revealed for the corresponding feature of the AI model. 

To accomplish the second task type, we can leverage SAT score characteristics as shown in the top row of Fig.~\ref{fig:variable_cdf}, focusing on the correctly classified test data. Features \textit{mean concave points} and \textit{worst concave points} are highly important according to LIME and SHAP, but their SAT score distribution characteristics show large probability masses in the range of very small percentage changes needed to invert the model prediction. The physician and the engineer from our examples should be very careful and, if possible, measure values of these features with very high accuracy. 
On the other hand, features that would need greater changes to modify their predictions such as \textit{worst concavity}, leaves our engineer with some margin of error, because non-trivial changes of its value are required to impact the model prediction. The engineer can compare this margin with the characteristics of various sensors that can be used to measure the particular property of interest, and choose the least expensive sensor whose error model safely fits within the identified slack of robustness of the AI model. 

Concentrated distributions show that the same magnitude change is present among a group of explanations for multiple data points. Some features may cause more fragility in the system than others because their percent change concentrates in the low range of values i.e. as observed in \textit{mean concave points}, while for \textit{worst concavity} a large change in the observations would be needed to break robustness thus the latter is more reliable. If the misclassified points have a percent change smaller than the rest of the points, it would mean that changing the features would most likely result in the improvement of the model. While this trend is prevalent in our dataset, a few counterexamples exist for which changing their values would most likely decrease the performance of the model.

%In our analysis of the stochasticity of LIME and SHAP, the feature importance graphs show there exist variable-to-variable interactions since the functions are not monotonically increasing, thus an important feature at $n$ is not guaranteed to also be important at $n+1$ demonstrating predictive multiplicity, or multiple ways to explain a certain phenomenon. This behavior can be observed in the bottom row of Figure ~\ref{fig:variable_cdf}. Multiple explanations are unstable and can cause a lack of trustworthiness in high-stakes applications. 

%

\section{Conclusion}
Model-centric formal methods provide useful capabilities complementary to the existing explanatory analysis tools.
They are based on mathematical logic and yield provable results that can be verified exactly, as opposite to the prevalent statistical methods that produce results with margins of confidence. We envision beneficial use of these methods at all stages of life of AI systems: from design to field application.

\section{Limitations and Future Work}
A limitation of the current work is that the implemented verification framework does not take into account the mutability of feature. Some features, such as age, cannot be changed, making the counterfactuals less actionable. This is a limitation of the implementation alone, and previous work has addressed this by making certain features protected ~\cite{gisolfi2021fairness}. In future work we hope to further validate these methods by applying them to a variety of datasets. Additionally, these methods should be validated with human studies, similar to those found in ~\cite{lime}.

\small % However, in this section (only), you may reduce the size to \textbackslash small if your paper exceeds the allowable number of pages.
\bibliography{aaai22.bib}

\begin{thebibliography}{16}
\providecommand{\natexlab}[1]{#1}
\providecommand{\url}[1]{\texttt{#1}}
\providecommand{\urlprefix}{URL }
\expandafter\ifx\csname urlstyle\endcsname\relax
  \providecommand{\doi}[1]{doi:\discretionary{}{}{}#1}\else
  \providecommand{\doi}{doi:\discretionary{}{}{}\begingroup
  \urlstyle{rm}\Url}\fi

\bibitem[{Alvarez-Melis and Jaakkola(2018)}]{AlvarezMelis2018OnTR}
Alvarez-Melis, D.; and Jaakkola, T. 2018.
\newblock On the Robustness of Interpretability Methods.
\newblock \emph{ArXiv} abs/1806.08049.

\bibitem[{Dua and Graff(2017)}]{Dua:2019}
Dua, D.; and Graff, C. 2017.
\newblock {UCI} Machine Learning Repository.
\newblock \urlprefix\url{http://archive.ics.uci.edu/ml}.

\bibitem[{Ghorbani, Abid, and Zou(2019)}]{ghorbani2019interpretation}
Ghorbani, A.; Abid, A.; and Zou, J. 2019.
\newblock Interpretation of neural networks is fragile.
\newblock In \emph{Proceedings of the AAAI Conference on Artificial
  Intelligence}, volume~33, 3681--3688.

\bibitem[{Gisolfi et~al.(2021{\natexlab{a}})Gisolfi, Good, De-Arteaga, Edman,
  and Dubrawski}]{gisolfi2021fairness}
Gisolfi, N.; Good, J.~H.; De-Arteaga, M.; Edman, R.; and Dubrawski, A.
  2021{\natexlab{a}}.
\newblock Formal Verification of Individual Fairness.
\newblock \emph{Under Review} .

\bibitem[{Gisolfi et~al.(2021{\natexlab{b}})Gisolfi, Good, Miller, and
  Dubrawski}]{gisolfi2021formal}
Gisolfi, N.; Good, J.~H.; Miller, K.; and Dubrawski, A. 2021{\natexlab{b}}.
\newblock Formal Verification of Voting Tree Ensembles.
\newblock \emph{Under Review} .

\bibitem[{Karimi et~al.(2020)Karimi, Barthe, Balle, and
  Valera}]{karimi2020model}
Karimi, A.-H.; Barthe, G.; Balle, B.; and Valera, I. 2020.
\newblock Model-agnostic counterfactual explanations for consequential
  decisions.
\newblock In \emph{International Conference on Artificial Intelligence and
  Statistics}, 895--905. PMLR.

\bibitem[{Lipton(2018)}]{10.1145/3236386.3241340}
Lipton, Z.~C. 2018.
\newblock The Mythos of Model Interpretability: In Machine Learning, the
  Concept of Interpretability is Both Important and Slippery.
\newblock \emph{Queue} 16(3): 31–57.
\newblock ISSN 1542-7730.
\newblock \doi{10.1145/3236386.3241340}.
\newblock \urlprefix\url{https://doi.org/10.1145/3236386.3241340}.

\bibitem[{Lundberg and Lee(2017)}]{NIPS2017_7062}
Lundberg, S.~M.; and Lee, S.-I. 2017.
\newblock A Unified Approach to Interpreting Model Predictions.
\newblock 4765--4774. Curran Associates, Inc.
\newblock
  \urlprefix\url{http://papers.nips.cc/paper/7062-a-unified-approach-to-interpreting-model-predictions.pdf}.

\bibitem[{Lundberg et~al.(2018)Lundberg, Nair, Vavilala, Horibe, Eisses, Adams,
  Liston, Low, Newman, Kim et~al.}]{lundberg2018explainable}
Lundberg, S.~M.; Nair, B.; Vavilala, M.~S.; Horibe, M.; Eisses, M.~J.; Adams,
  T.; Liston, D.~E.; Low, D. K.-W.; Newman, S.-F.; Kim, J.; et~al. 2018.
\newblock Explainable machine-learning predictions for the prevention of
  hypoxaemia during surgery.
\newblock \emph{Nature Biomedical Engineering} 2(10): 749.

\bibitem[{Molnar(2019)}]{molnar2019}
Molnar, C. 2019.
\newblock \emph{Interpretable Machine Learning}.
\newblock \url{https://christophm.github.io/interpretable-ml-book/}.

\bibitem[{Morales et~al.(2019)Morales, Carter, Tan, and
  Steinfeld}]{10.1145/3322276.3322345}
Morales, C.~G.; Carter, E.~J.; Tan, X.~Z.; and Steinfeld, A. 2019.
\newblock Interaction Needs and Opportunities for Failing Robots.
\newblock In \emph{Proceedings of the 2019 on Designing Interactive Systems
  Conference}, DIS '19, 659–670. New York, NY, USA: Association for Computing
  Machinery.
\newblock ISBN 9781450358507.
\newblock \doi{10.1145/3322276.3322345}.
\newblock \urlprefix\url{https://doi.org/10.1145/3322276.3322345}.

\bibitem[{Narodytska et~al.(2019)Narodytska, Shrotri, Meel, Ignatiev, and
  Marques-Silva}]{narodytska2019assessing}
Narodytska, N.; Shrotri, A.; Meel, K.~S.; Ignatiev, A.; and Marques-Silva, J.
  2019.
\newblock Assessing heuristic machine learning explanations with model
  counting.
\newblock In \emph{International Conference on Theory and Applications of
  Satisfiability Testing}, 267--278. Springer.

\bibitem[{Pedregosa et~al.(2011)}]{scikit-learn}
Pedregosa, F.; et~al. 2011.
\newblock Scikit-learn: Machine Learning in {P}ython.
\newblock \emph{Journal of Machine Learning Research} 12: 2825--2830.

\bibitem[{Reig et~al.(2018)Reig, Norman, Morales, Das, Steinfeld, and
  Forlizzi}]{10.1145/3239060.3239064}
Reig, S.; Norman, S.; Morales, C.~G.; Das, S.; Steinfeld, A.; and Forlizzi, J.
  2018.
\newblock A Field Study of Pedestrians and Autonomous Vehicles.
\newblock In \emph{Proceedings of the 10th International Conference on
  Automotive User Interfaces and Interactive Vehicular Applications},
  AutomotiveUI '18, 198–209. New York, NY, USA: Association for Computing
  Machinery.
\newblock ISBN 9781450359467.
\newblock \doi{10.1145/3239060.3239064}.
\newblock \urlprefix\url{https://doi.org/10.1145/3239060.3239064}.

\bibitem[{Ribeiro, Singh, and Guestrin(2016)}]{lime}
Ribeiro, M.~T.; Singh, S.; and Guestrin, C. 2016.
\newblock "Why Should I Trust You?": Explaining the Predictions of Any
  Classifier.
\newblock In \emph{Proceedings of the 22nd ACM SIGKDD International Conference
  on Knowledge Discovery and Data Mining}, 1135–1144. New York, NY, USA:
  Association for Computing Machinery.
\newblock ISBN 9781450342322.
\newblock \doi{10.1145/2939672.2939778}.
\newblock \urlprefix\url{https://doi.org/10.1145/2939672.2939778}.

\bibitem[{Saarela and Jauhiainen(2021)}]{article}
Saarela, M.; and Jauhiainen, S. 2021.
\newblock Comparison of feature importance measures as explanations for
  classification models.
\newblock \emph{SN Applied Sciences} 3.
\newblock \doi{10.1007/s42452-021-04148-9}.

\end{thebibliography}

\newpage
\onecolumn 
\section{Appendix}

\begin{figure}[!h]
    \centering
    \includegraphics[width=\textwidth, height=20cm]{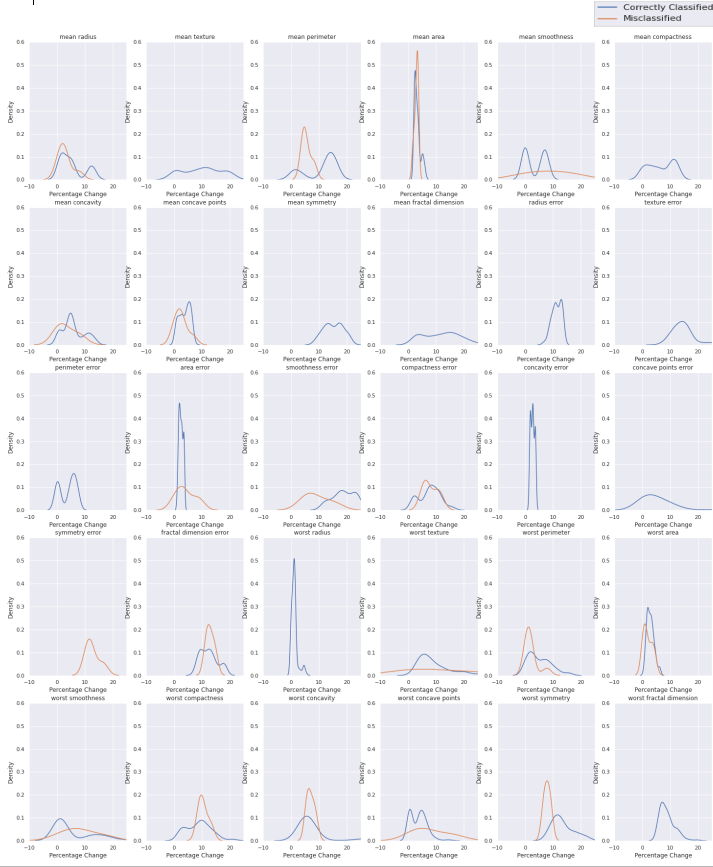}
    \caption{Aggregate summary of SAT counterfactual explanations }
    \label{fig:sat_summary}
\end{figure}
\newpage
\begin{figure*}
    \centering
    \includegraphics[width=\textwidth, height=22cm]{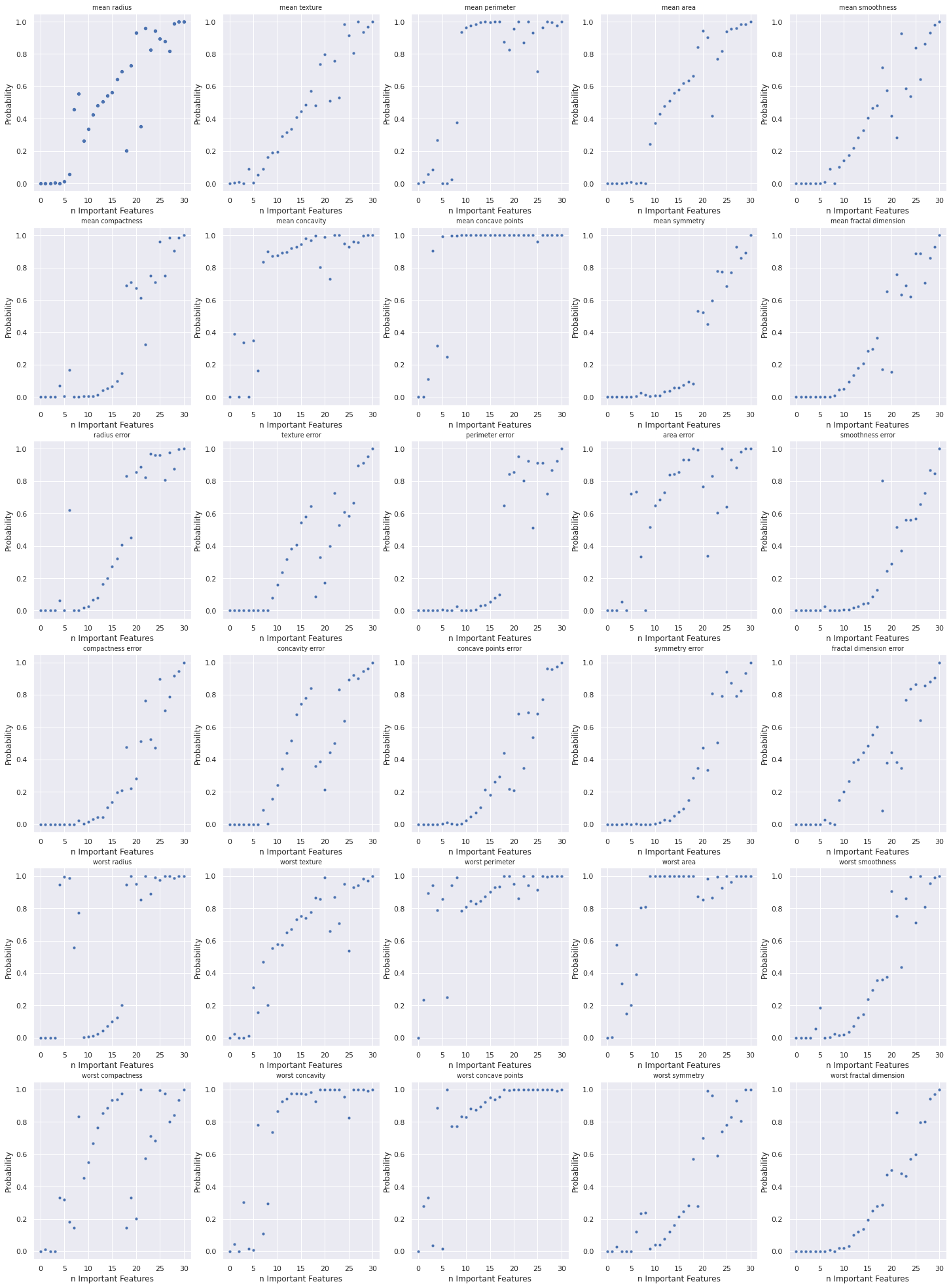}
    \caption{Probability of feature importance in LIME when n features are chosen.}
    \label{fig:lime_summary}
\end{figure*}

\begin{figure*}
    \centering
    \includegraphics[width=\textwidth, height=22cm]{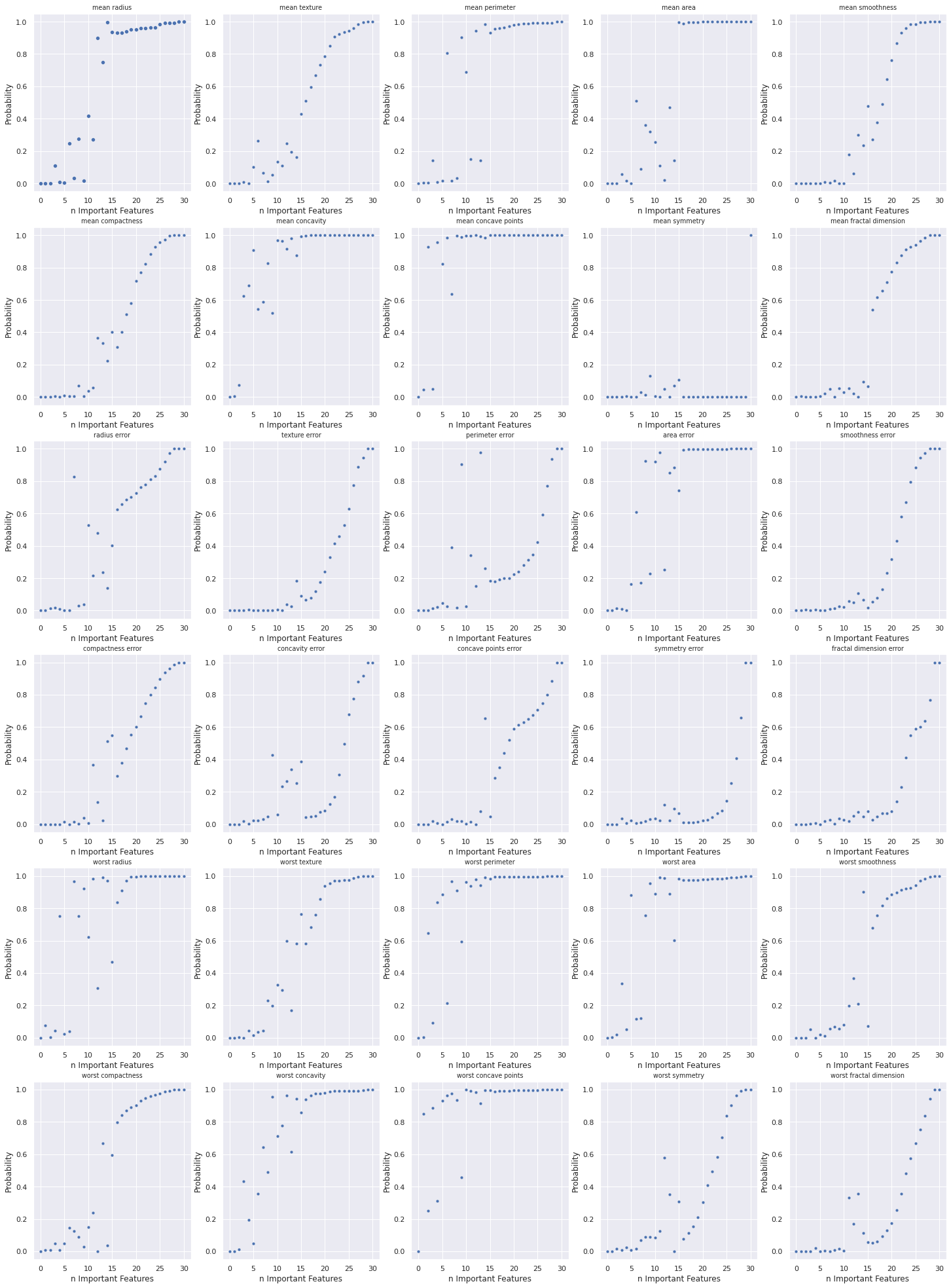}
    \caption{Probability of feature importance in SHAP when n features are chosen.}
    \label{fig:shap_summary}
\end{figure*}

\end{document}